\pdfoutput=1

\documentclass[11pt]{article}

\newcommand{\Furina}{F{\small URINA}}

\usepackage[preprint]{acl}

\usepackage{times}
\usepackage{latexsym}
\usepackage{amssymb}

\usepackage{bbding}
\usepackage{colortbl}

\usepackage{graphicx}
\usepackage{amsmath}
\usepackage{booktabs}
\usepackage{arydshln}
\usepackage{xcolor}
\usepackage{multirow}

\usepackage[utf8]{inputenc}

\usepackage{microtype}

\usepackage{inconsolata}

\usepackage[T1]{fontenc}

\usepackage[utf8]{inputenc}

\usepackage{microtype}

\usepackage{inconsolata}

 \usepackage{colortbl}

\title{MaiNLP at SemEval-2024 Task 1: Analyzing \\Source Language Selection in Cross-Lingual Textual Relatedness}

\author{Shijia Zhou$^{1,*}$ \quad
Huangyan Shan$^{1,}$\thanks{\quad Both authors contributed equally.} \quad 
\textbf{Barbara Plank}$^{1,2}$ \quad 
Robert Litschko$^{1,2}$\\ 
  \textsuperscript{1}MaiNLP, Center for Information and Language Processing, LMU Munich, Germany \\
    \textsuperscript{2}Munich Center for Machine Learning (MCML), Munich, Germany \\
{\tt \{zhou.shijia, Shan.Huangyan\}@campus.lmu.de} \hspace{1em} 
{\tt \{bplank, rlitschk\}@cis.lmu.de}}

\begin{document}
\maketitle
\begin{abstract}
This paper presents our system developed for the SemEval-2024 Task 1: Semantic Textual Relatedness (STR), on Track C: Cross-lingual. The task aims to detect semantic relatedness of two sentences in a given target language without access to direct supervision (i.e. zero-shot cross-lingual transfer). %
To this end, we focus on different source language selection strategies on two different pre-trained languages models: XLM-R and \Furina{}. 
We experiment with 1) single-source transfer and select source languages based on typological similarity, 2) augmenting English training data with the two nearest-neighbor source languages, and 3) multi-source transfer where we compare selecting on all training languages against languages from the same family. 
We further study machine translation-based data augmentation and the impact of script differences. 
Our submission achieved the first place in the C8 (Kinyarwanda) test set.
\end{abstract}

\section{Introduction}
The task of semantic textual relatedness (STR) has a long-standing tradition in NLP~\cite[e.g.,][]{mohammad2008measuring}. It consists of predicting a score that reflects the closeness in semantic meaning between two given sentences. For example, consider the following examples extracted from the actual shared task data~\cite{abdalla-etal-2023-makes} shown in Figure~\ref{fig:examples}. For English, the annotators scored the first pair higher than the second sentence pair. Similarly, for Afrikaans the annotators scored the first example higher than the second one. As further described in~\citet{abdalla-etal-2023-makes}, all sentence pairs were annotated manually in a pairwise fashion to obtain semantic textual relatedness (STR) scores between 0 (completely unrelated) and 1 (maximally related).

\begin{figure}
\begin{small}
    \centering
    \begin{tabular}{p{0.5cm}p{0.4cm}p{5.5cm}}
    \textbf{Pair} & \textbf{STR} & \textbf{Sentence Pair}\\
    \toprule
    \multirow{ 2}{*}{eng-25} & \multirow{ 2}{*}{0.88} & ``It is better known as a walk.''\\
    & & 
``It is also known as a walk .'' \\
\midrule
    \multirow{ 2}{*}{eng-31}  & \multirow{ 2}{*}{0.30}  & ``But, of course, it's not that simple'' \\
    & & ``However, this is not for me.''\\
\midrule
    \multirow{ 2}{*}{afr-87}  & \multirow{ 2}{*}{0.72}  &  ``ols totdat dit n bal vorm.''\\
    & & ``Dit moet n stywe bal deeg vorm.''\\
\midrule
    \multirow{ 4}{*}{afr-78}  & \multirow{ 4}{*}{0.09}  &  ``Stel jou voor jou kind skryf elke week n opstel.'' \\ 
    & &  ``Washington is ook n fietsryer-vriendelike stad.''\\ %
    \bottomrule
    \end{tabular}
    \end{small}
    \caption{Examples from the dev sets for Semantic Textual Relatedness (STR). eng: English, afr: Afrikaans. }
    \label{fig:examples}
\end{figure}

While previous work has largely focused on English, the
SemEval-2024 shared task 1~\citep{semrel2024task} aims to extend the language coverage. 
It proposes datasets to evaluate the relatedness of sentence pairs for a total of 14 languages, including low-resource tail languages such as Kinyarwanda (kin) or Marathi (mar) \citep{abdalla-etal-2023-makes} (see \S\ref{sec:datasets}). The shared task includes three subtracks, each with a focus on supervised, unsupervised and cross-lingual STR, respectively. In this paper, we focus on Track C, \textit{cross-lingual STR}. In this track, the goal is to develop a system to predict STR scores  \textit{without} access to any labeled data for the target language (importantly, also no target development data). 
That is, Track C requires the development of a regression model for 12 target languages, without relying on any labeled datasets in the target language (or pre-trained language model fine-tuned on other STR tasks). 
Instead the cross-lingual task allows to utilize training datasets from at least one other language from the other tracks (which includes training data of up to 9 languages). Returning to our running example in Figure~\ref{fig:examples}, the task is to develop a system for example for Afrikaans as target by transferring knowledge from one or more source languages (which may include English).

Previous work on multilingual NLP has illustrated the \textit{curse of multilinguality}~\cite{conneau-etal-2020-unsupervised}, that is, diminishing returns for training a single system on many languages due to language interference. %
This shared task has a focus on low-resource languages and languages typologically distant to English, a setup in where cross-lingual transfer has shown to be particularly challenging \citep{lauscher-etal-2020-zero}. 
Motivated by these two aspects, we set out to study the use of fewer but more relevant \textit{source} languages for a given target language. More specifically, we aim to find good ``donor language(s)''~\cite{malkin-etal-2022-balanced} and compare those to baselines that either only use English, or a multi-source model trained on all source languages (except the target). We aim to answer the following research questions: \textbf{RQ1} To what extent does knowledge transfer from source languages improve STR models? \textbf{RQ2} Do multilingual STR models exhibit language interference \citep{wang-etal-2020-negative}, i.e., performance drops when training data from heterogeneous languages are combined? \textbf{RQ3} To what extent do script differences play a role in STR (``script gap''), and can we narrow the script gap by using a foundation model specialized to align transliterated data and data written in different scripts? \textbf{RQ4} Can we further improve the transfer performance by relying on machine translation to augment existing training data?

To study RQ1, we make use of typological information available in language vectors. For RQ2, we opt for a multi-source approach, that combines the training data for all languages (except the target). To study the impact of scripts (RQ3), we make use of transliteration, and further compare a BERT-based model to \Furina~\citep{liu2024translico}, a recently proposed language model that aims to better align languages across scripts. Finally for RQ4, we investigate the use of machine translation (MT) for data augmentation.
We apply ours methods to 12 target languages in Track C. The  specific details about languages are presented in \S\ref{sec:datasets}.

\setlength{\tabcolsep}{5.7pt}
\begin{table*}[t!]
\centering
\small 
\begin{tabular}{l c c c c c c c c c c c c c c c}
\toprule
 & eng & esp & afr & hin & pan & amh & arb & arq & ary & hau & ind & kin & mar & tel & total \\ \midrule
Train & 5,500 & 1,562 & - & - & - & 992 & - & 1,261 & 924 & 1,736 & - & 778 & 1,200 & 1,170 & 15,123 \\
Dev & 249 & 139 & 375 & 288 & 242 & 95 & 32 & 97 & 70 & 212 & 144 & 222 & 293 & 130 & 2,588 \\
Test & 2,600 & 140 & 375 & 968 & 634 & 171 & 595 & 583 & 425 & 594 & 360 & 222 & - & - & 7,667 \\
\bottomrule
\end{tabular}
\caption{STR Dataset statistics. Indo-European lanuguages including esp, afr, hin, ind, pan and mar: 10,424 train instances; 1,811 dev instances; 5,357 test instances. Afro-Asiatic languages including hau, amh, arb, ary and arq: 3,921 train instances; 411 dev instances; 2,197 test instances. Out of 14 languages, 5 languages including amh, hin, arb, arq, ary are in non-latin script, all the rest of languages are in latin script. %
}
\label{tab:stats}
\end{table*}

\section{Background}
\subsection{STR Task Setup and Datasets} \label{sec:datasets}
The STR task \citep{semrel2024task} aims to measure the extent to which two linguistic elements share semantic proximity \citep{semrel2024dataset}. 
These elements may be associated through various means, such as conveying similar ideas, originating from the same historical period, complementing each other's meaning, and so forth. 
It offers 3 tracks to follow: supervised (Track A), unsupervised (Track B), cross-lingual (Track C). 
In Track C, participants must provide systems developed without relying on any labeled datasets specifically tailored for semantic similarity or relatedness in the target language. 
Instead, they are required to employ labeled dataset(s) from at least one other language. 

The STR task involves 14 monolingual datasets for Afrikaans (\texttt{afr}), Amharic (\texttt{amh}), Modern Standard Arabic (\texttt{arb}), Algerian Arabic (\texttt{arq}), Moroccan Arabic (\texttt{ary}), English (\texttt{eng}), Spanish (\texttt{esp}), Hausa (\texttt{hau}), Hindi (\texttt{hin}), Indonesian (\texttt{ind}), Kinyarwanda (\texttt{kin}), Marathi (\texttt{mar}), Punjabi (\texttt{pan}), and Telugu (\texttt{tel}). 
Among these, Track~A and Track~C comprise 9 and 12 languages respectively (see Table~\ref{tab:stats}).
In the training datasets, each instance consists of a sentence pair and is assigned a golden STR score as judged by native speakers. 
The score ranges between 0 and 1, with higher values indicating greater relatedness between the sentence pairs. For details on the data collection, we refer the reader to the shared task overview paper~\cite{semrel2024dataset}.

As per requirement, we designate the 9 languages in Track A as source languages and those in Track C as the 12 target languages. An overview of the resulting train/dev/test data statistics for the 14 languages is provided in Table~\ref{tab:stats}.

\subsection{Evaluation Metric}
The evaluation metric used in  this shared task is Spearman's rank correlation coefficient. It  evaluates the strength and direction of the monotonic relationship between two variables with a range from -1 to 1. In the context of our task, as previously mentioned, the scoring has been adjusted to range between 0 and 1. We use the evaluation script provided by the organizers.\footnote{\url{https://github.com/semantic-textual-relatedness/Semantic_Relatedness_SemEval2024/blob/main/semrel_baselines/src/evaluate_sbert.py}}

\subsection{Baselines}
The organizers fine-tuned LaBSE \citep{feng-etal-2022-language} on the English training set to get baselines for all target languages except English (cf. §\ref{subsec:modelselection}). For English, they fine-tuned LaBSE on Spanish as a baseline. Since the test dataset for Spanish has not been made publicly available, all models aimed at Spanish evaluation are conducted solely on their respective validation datasets. In order to ensure a more equitable comparison with other findings, we reproduce the baseline LaBSE model utilizing the methodology provided by the organizers. It yields a baseline score of 0.687 on the Spanish validation dataset.

\section{Methods}

We opt for two RoBERTa-based \citep{liu2019roberta} models for the regression task trained with a mean-squared error (MSE) loss. More specifically, we use the XLM-RoBERTa base model, and \Furina~\cite{liu2024translico}, which is a XLM-R derivative based on Glot-500 \citep{imanigooghari-etal-2023-glot500}, further detailed below. %
We adopt a multi-source approach that involves individually fine-tuning a model for each target language in Track~C. This fine-tuning process utilizes the training datasets from all languages available in Track A, explicitly excluding the dataset of the test language itself. For baseline comparisons, we use XLM-RoBERTa \citep{conneau-etal-2020-unsupervised} and \Furina~\cite{liu2024translico} models fine-tuned solely on English datasets.

\subsection{Model Selection}
\label{subsec:modelselection}

\paragraph{XLM-RoBERTa} The multilingual masked language model XLM-RoBERTa (XLM-R) \citep{conneau-etal-2020-unsupervised} pre-trained on 2.5TB of filtered CommonCrawl data containing 100 languages has shown superior performance compared to Multilingual BERT (mBERT) \citep{devlin-etal-2019-bert} across a range of cross-lingual benchmarks. In the experiment, we utilize the base version of XLM-R.\footnote{\url{https://huggingface.co/FacebookAI/xlm-roberta-base}} XLM-R has seen all SemRelEval languages except for Algerian Arabic (arq), Moroccan Arabic (ary), Kinyarwanda (kin) at pre-training time.

\paragraph{\Furina} \Furina~\citep{liu2024translico} covers 511 low-resource languages. It was fine-tuned on Glot500-m \citep{imanigooghari-etal-2023-glot500}. The training data consists of 5\% of Glot500-m's pretraining sentences in original script as well as their corresponding Latin transliterations. At pre-training time \Furina{} has been exposed to all SemRelEval languages except for Algerian Arabic (arq).

\paragraph{LaBSE} The shared task organizers provide cross-lingual baselines for each target language by fine-tuning Label Agnostic BERT Sentence Embeddings (\texttt{LaBSE}) \citep{feng-etal-2022-language}, which supports 109 languages. LaBSE was pre-trained using Translation language modeling (TLM) \citep{lample2019cross}, which included bilingual translation sentence pairs for training. 
The bilingual corpus is constructed from web pages using a bitext mining system, filtered by a pre-trained contrastive data-selection scoring model, and manually curated to create a high-quality collection of 6 billion translation pairs. %
Out of those, LaBSE has been exposed to different amounts of parallel data (\texttt{eng}-\texttt{xxx}) from SemRelEval languages. The largest amount of parallel text involves Spanish with over 375M sentence pairs (\texttt{eng}-\texttt{esp}), followed by Indonesian with over 250M sentence pairs (eng-ind), followed by Hindi and Arabic (\texttt{eng}-\{\texttt{hin}, \texttt{arb}\}) with over 125M language pairs. All other languages (\texttt{afr}, \texttt{pan}, \texttt{amh}, \texttt{haus}, \texttt{tel}, \texttt{kin}, \texttt{mar}) appear in the TLM training corpus with less than 125M sentence pairs.

\subsection{Source Language Selection}
\label{sec:da}

\paragraph{Single-Source Transfer} 
In our first approach, we follow the standard single-source zero-shot cross-lingual transfer setup and fine-tune pre-trained language models on English data ($\texttt{XLM-R}_\texttt{eng}$, $\texttt{Furina}_\texttt{eng}$). This is a common evaluation approach adopted in standard natural language understanding and generation benchmarks \citep{liang-etal-2020-xglue,ruder-etal-2023-xtreme}. 
However, English has been shown to not always be the best source language \citep{turc2021revisiting}.
To investigate if this also true for SemRelEval, we further experiment with selecting for each test language its closest (i.e., most similar) source language. Here, we measure language similarity according to typological features from the lang2vec library \citep{littell-etal-2017-uriel}.

\paragraph{K-nearest-neighbor languages} %
In this approach we augment the English training dataset with the datasets of $k$ languages that are closest to the target language, dubbed \texttt{kNN}. To determine suitable source languages for each target language, we assess language similarity by calculating the cosine similarity between language vectors learned by a multilingual neural MT model provided by \citet{malaviya-etal-2017-learning}. 
We specifically use the cell\_state language vectors, which are computed by encoding all sentences in a given language and then computing the average hidden cell state of the encoder LSTM.\footnote{\url{https://github.com/chaitanyamalaviya/lang-reps/}}
These vectors can be seen as language embeddings encoding latent typology features \citep{ostling-tiedemann-2017-continuous,yu-etal-2021-language}. With our \texttt{kNN}-models we aim for a good balance between large amounts of training instances (English) and positive transfer from similar languages.

\paragraph{Multi-Source Transfer}
The STR dataset contains languages from different language families. To investigate whether training a single model on a disverse set of languages leads to negative interference \citep{wang-etal-2020-negative} we compare two multi-source models. In the first model, dubbed \texttt{MS-All}, we fine-tune XLM-R and Furina on the concatenation of all training sets from Track A (excluding the target language). Inspired by previous work on combining \textit{multiple related} source languages \citep{snaebjarnarson-etal-2023-transfer,lim2024analysis}, we further evaluate multi-source models trained on languages from the same language family (\texttt{MS-Fam}).

\subsection{Other Approaches}

\paragraph{Machine Translation}

For the purpose of data augmentation and balance of languages, we translate selected languages into each other using NLLB \citep{nllbteam2022language}, ensuring that each language contributes equally to the training dataset. Taking Kinyarwanda as an example, we select Hausa and Spanish as the two languages closest to it, based on dense language vector similarity as outlined above (\texttt{kNN}), along with English, as training dataset. We translate among these three languages mutually, thus tripling the size of the training dataset while ensuring a balanced representation of all languages.

\paragraph{Transliteration}

Additionally, we attempt to further facilitate multilingual transfer learning by standardizing script across languages. Utilizing the tool Uroman\footnote{\url{https://github.com/isi-nlp/uroman}} \citep{hermjakob-etal-2018-box}, which was also used by \Furina~\citep{liu2024translico}, we transliterate the train and test datasets of languages written in non-Latin scripts, including both the original datasets and the translated datasets, into Latin script. 
We evaluate the models fine-tuned on Romanized training data on the Romanized test dataset.
This attempt only involves non-Latin script languages (\texttt{amh}, \texttt{arb}, \texttt{ary}, \texttt{arq}, \texttt{hin}).%

\setlength{\tabcolsep}{6.5pt}
\begin{table*}%
\centering
\small 
\begin{tabular}{l c c c c c c c c c c c c c}
\toprule
& \multicolumn{5}{c}{Indo-European} & \multicolumn{5}{c}{Afro-Asiatic} & \multicolumn{2}{c}{Other} \\ \cmidrule(lr){2-6} \cmidrule(lr){7-11} \cmidrule(lr){12-13}
 & eng & esp & afr & hin & pan & amh & arb & arq & ary & hau & ind & kin & avg \\
\midrule 
\texttt{LaBSE} (baseline) & 0.80 & 0.69 & 0.79 & 0.76 & -0.05 & \textbf{0.84} & \textbf{0.61} & 0.46 & 0.40 & 0.62 & \textbf{0.47} & 0.57 & 0.67 \\
$\texttt{Furina}_\texttt{eng+esp+hau}$ & - & - & 0.74 & 0.70 & \textbf{0.09} & 0.73 & 0.40 & 0.27 & 0.57 & - & 0.32 & 0.68 & - \\
\midrule
\multicolumn{14}{c}{\textit{Models based on XLM-R} \citep{conneau-etal-2020-unsupervised}}  \\ \midrule
$\texttt{XLM-R}_{\texttt{eng}}$ & - & 0.67 & \textbf{0.81} & 0.80 & -0.02 & 0.81 & 0.60 & 0.50 & 0.60 & 0.64 & 0.42 & 0.46 & 0.71 \\ \cdashline{1-14}[.4pt/1pt]\noalign{\vskip 0.51ex}
$\texttt{XLM-R}_\texttt{MS-All}$ & \textbf{0.84}  & 0.63 & 0.80 & \textbf{0.82} & -0.01 & 0.80 & 0.56 & 0.59 & 0.82 & 0.66 & 0.42 & 0.69 & \textbf{0.73} \\
$\texttt{XLM-R}_\texttt{MS-Fam}$ & 0.82 & 0.71 & \textbf{0.81} & \textbf{0.82} & 0.00 & 0.69 & 0.44 & 0.37 & \textbf{0.83} & 0.66 & - & - & 0.68 \\ \cdashline{1-14}[.4pt/1pt]\noalign{\vskip 0.51ex}
$\texttt{XLM-R}_\texttt{kNN}$ & - & 0.59 & \textbf{0.81} & 0.78 & - & 0.75 & 0.57 & - & 0.50 & 0.62 & 0.45 & 0.41 & 0.69 \\ 
$\texttt{XLM-R}_\texttt{kNN+MT}$ & - & 0.64 & 0.80 & 0.78 & \ - & 0.77 & 0.54 & - & 0.55 & 0.62 & 0.36 & 0.55 & 0.70 \\
$\texttt{XLM-R}_\texttt{kNN+TL}$ & - & - & - & 0.66 & - & 0.37 & 0.45 & - & 0.52 & - & -  & - & -  \\
\midrule
\multicolumn{14}{c}{\textit{Models based on Furina} \citep{liu2024translico}}  \\ \midrule
$\texttt{Furina}_\texttt{eng}$ & - & 0.54 & 0.79 & 0.70 & -0.14 & 0.74 & 0.37 & 0.45 & 0.59 & 0.63 & 0.44 & 0.53 & 0.62 \\ \cdashline{1-14}[.4pt/1pt]\noalign{\vskip 0.51ex}
$\texttt{Furina}_\texttt{MS-All}$ & 0.83 & 0.59 & 0.79 & 0.76 & -0.02 & 0.81 & 0.49 & \textbf{0.61} & \textbf{0.83} & 0.65 & 0.35 & \textbf{0.78} & 0.71 \\
$\texttt{Furina}_\texttt{MS-Fam}$ & 0.83 & \textbf{0.72} & 0.79 & 0.77 & 0.02 & 0.66 & 0.42 & 0.55 & 0.82 & \textbf{0.68} & - & - & 0.71 \\ \cdashline{1-14}[.4pt/1pt]\noalign{\vskip 0.51ex}
$\texttt{Furina}_\texttt{kNN}$ & - & 0.59 & 0.80 & 0.72 & - & 0.74 & 0.43 & - & 0.57 & 0.63 & 0.46 & 0.68 & 0.67 \\ 
$\texttt{Furina}_\texttt{kNN+MT}$ & - & 0.56 & 0.78 & 0.75 & - & 0.74 & 0.44 & - & 0.57 & 0.59 & 0.37 & 0.64 & 0.67 \\ 
$\texttt{Furina}_\texttt{kNN+TL}$ & - & - & - & 0.67 & - & 0.72 & 0.44 & - & 0.56 & - & -  & - & -  \\
\bottomrule
\end{tabular}

\caption{Spearman's rank correlation of zero-shot transfer experiments on SemRelEval 9 test languages. The organizers decided to keep the test set for Spanish private, we therefore report the performance on the validation set. We exclude English from the average result (avg). \textbf{bold}: Best result for each language. Languages not covered by all L2V features are excluded from the average (\texttt{eng}, \texttt{pan}, \texttt{arq}, \texttt{ind}, \texttt{kin}). For our \texttt{kNN}-variants we opt for $k=2$.}
\label{tab:results}
\end{table*}

\section{Experimental Setup}

The detailed setting are listed in Appendix \ref{sec:Hyperparameter}.
As baseline, we exclusively train a  model on the English dataset ($\texttt{XLM-R}_\texttt{eng}$, $\texttt{Furina}_\texttt{eng}$) and assess its performance across all target languages.
Subsequently, for each target language, we fine-tune a multi-source model: if the target language is not within the 9 training datasets, we train on the union of all $n=9$ training languages. Otherwise we train a multi-source model on $n-1=8$ source languages, excluding the target ($\texttt{XLM-R}_\texttt{MS-All}$, $\texttt{Furina}_\texttt{MS-All}$).
Following this, we explore whether it is helpful to prune certain languages, retaining only English and the two closest to the target languages according to lang2vec~\cite{littell-etal-2017-uriel}\footnote{We compare the similarity of languages based on three criteria: lang\_cell\_states, lang\_vecs and language typological vectors} as source languages ($\texttt{XLM-R}_\texttt{L2V}$, $\texttt{Furina}_\texttt{L2V}$). Due to the reduction in the training set, which significantly decreased the size of the data, we attempted to expand the dataset through cross-translation ($\texttt{XLM-R}_\texttt{L2V-Aug}$, $\texttt{Furina}_\texttt{L2V-Aug}$;
cf.~§\ref{sec:da}).

\section{Results and Discussion}

Our main results are presented in Table~\ref{tab:results} and are discussed in the following section.

\paragraph{Single-source versus multi-source transfer.}
We first compare the performance of a zero-shot STR model trained on English ($\texttt{XLM-R}_\texttt{eng}$, $\texttt{Furina}_\texttt{eng}$) against a multi-source model trained on the concatenation of all available languages from Track~A ($\texttt{XLM-R}_\texttt{MS-All}$, $\texttt{Furina}_\texttt{MS-All}$). Our results reveal that knowledge transfer from multiple source languages (\textbf{RQ1}) improves STR models, %
affirming the potential of multi-source training to enhance cross-lingual capabilities. On average, both \texttt{MS-All} models outperform their single-source counterparts by 0.02 and 0.09 respectively. This is expected since the multi-source training dataset is with 15,123 instances almost three times larger than the English dataset with 5,500 instances (cf.\ Table~\ref{tab:stats}). %
When trained solely on English data, \Furina{} performs substantially worse than $\text{XLM-R}$. However, this performance gap narrows when transitioning from single-source to multi-source training. 

\paragraph{Transfer from language families.}
After showing that models trained on all languages outperform the single-source baseline, we now investigate the effect of training on languages from the same family as source languages. Here we experiment with two multi-source models specialized only on Indo-European and Afro-Asiatic languages respectively (\texttt{MS-Fam}). Importantly, for each target language we train a multi-source model on all other languages in the same language family.\footnote{Indonesian and Kinyarwanda are the only SemRel languages in their family, we therefore cannot evaluate multi-source for those languages.} On Indo-European languages, we find that $\texttt{XLM-R}_{\texttt{MS-Fam}}$ and $\texttt{Furina}_{\texttt{MS-Fam}}$ yield similar results with much less training data (i.e., 4,913 fewer instances belonging to other language families). For Spanish, our models show performance gains of +0.8 and +0.13 for XLM-R and \Furina{} respectively, when compared to models trained on all languages. This underscores the presence of language interference \citep{wang-etal-2020-negative} in multilingual STR models when the training data from dissimilar languages are combined (\textbf{RQ2}). On Afro-Asiatic languages, we observe average performance drops of -0.09 and -0.06 for XLM-R and \Furina{} when moving from \texttt{MS-All} to \texttt{MS-Fam}. We hypothesize that this can be attributed to the amount of training data available. In fact, there are ~28\% fewer training instances for all Afro-Asiatic languages than for English (5,500).

\paragraph{Transfer from nearest language neighbors.} We now investigate the transfer performance when training STR models on their two closest languages according to cosine similarity of language cell state vectors, i.e. learned language vectors presented in \cite{malaviya-etal-2017-learning}. As mentioned earlier, we add English due to its large scale as a third training language. 
Our submitted system, $\texttt{Furina}_\texttt{eng+esp+hau}$, is trained on the two closest training languages of Kinyarwanda (\texttt{kin}) and has been ranked first place on the shared task leaderboard. Applying the same approach for each test language ($\texttt{XLM-R}_\texttt{kNN}$, $\texttt{Furina}_\texttt{kNN}$) shows mixed results. This indicates that the strong performance on \texttt{kin} can be attributed to the fact that, contrary to XLM-R, \texttt{kin} has been seen by Furina during pretraining.

\setlength{\tabcolsep}{7.3pt}
\begin{table*}[t!]
\centering
\small 
\begin{tabular}{l c c c c c c c c c c c c l } \toprule
& \multicolumn{5}{c}{Indo-European} & \multicolumn{5}{c}{Afro-Asiatic} & \multicolumn{2}{c}{Other} \\ \cmidrule(lr){2-6} \cmidrule(lr){7-11} \cmidrule(lr){12-13}
\textbf{XLM-R}
& eng & esp & afr & hin & pan & amh & arb & arq & ary & hau & ind & kin & avg \\ \midrule
M{\scriptsize IN} & 0.78 & 0.57 & 0.74 & 0.71 & -0.14 & 0.73 & 0.47 & 0.39 & 0.40 & 0.40 & 0.31 & 0.41 & 0.58 \\ \midrule
$\texttt{XLM-R}_\texttt{eng}$ & - & 0.67 & 0.81 & 0.80 & -0.02 & 0.81 & 0.60 & 0.50 & 0.60 & 0.64 & 0.42 & 0.46 & 0.71 \\ 
$\texttt{XLM-R}_\texttt{kNN}$ & - & 0.68 & 0.74 & 0.72 & - & 0.75 & 0.57 & - & 0.40 & 0.63 & 0.49 & 0.43 & 0.64 \\ \cdashline{1-14}[.4pt/1pt]\noalign{\vskip 0.51ex} 
\texttt{L2V-Pho} & 0.78 & 0.67 & 0.80 & 0.80 & -0.03 & 0.78 & 0.51 & 0.58 & 0.55 & 0.58 & 0.39 & 0.47 & 0.67 \\
\texttt{L2V-Syn} & 0.78 & 0.67 & 0.83 & 0.80 & -0.03 & 0.74 & 0.51 & 0.58 & 0.55 & 0.61 & 0.39 & 0.43 & 0.67 \\
\texttt{L2V-Inv} & 0.82 & 0.63 & 0.80 & 0.80 & -0.03 & 0.79 & 0.51 & 0.58 & 0.55 & 0.58 & 0.33 & 0.45 & 0.67 \\
\texttt{L2V-Fam} & 0.82 & 0.67 & 0.83 & 0.80 & -0.03 & 0.79 & 0.51 & 0.58 & 0.55 & 0.595 & - & - & 0.68 \\
\texttt{L2V-Geo} & 0.78 & 0.57 & 0.80 & 0.80 & -0.03 & 0.75 & 0.47 & 0.56 & 0.55 & 0.63 & 0.31 & 0.45 & 0.66 \\ %
\texttt{L2V-LRN} & - & 0.68 & 0.74 & 0.72 & - & 0.79 & 0.57 & - & 0.54 & 0.40 & 0.49 & 0.43 & 0.63 \\ \midrule
MAX & 0.82 & 0.69 & 0.83 & 0.80 & 0.04 & 0.79 & 0.61 & 0.63 & 0.74 & 0.66 & 0.49 & 0.65 & 0.73 \\  \bottomrule

\end{tabular}
\caption{
Single-source transfer results in terms of spearman correlation. The language selection is based on the cosine similarity of different typological features obtained from lang2vec (\texttt{L2V}). We additionally report the lower (MIN) and upper bound (MAX) obtained from selecting the best and worst source language. 
Languages not covered by all L2V features are excluded from the average: eng, pan, arq, ind, kin. 
For \texttt{L2V-Phon}, both tel and mar are closest to hin. For \texttt{L2V-Fam} amh and arq are the closest languages, we report their average score (0.595). 
In single-source transfer with  $\texttt{XLM-R}_\texttt{kNN}$ we use $k=1$ and do not combine the selected language with eng training data. 
}
\label{tab:results-l2v-ablation}
\end{table*}

\paragraph{Transliteration and cross-translation.} %
The STR dataset contains five test languages in non-Latin scripts: Hindi (\texttt{hin}), Amharic (\texttt{amh}), Standard Arabic (\texttt{arb}), Algerian Arabic (\texttt{arq}), and Moroccan Arabic (\texttt{ary}). %
Zero-shot cross-lingual transfer of models fine-tuned on English performs worse for Arabic scripts than for \texttt{amh} and \texttt{hin}. 
When fine-tuned on on multiple source languages (\texttt{MS-All}) XLM-R improves the performance on three out of five languages while Furina yields improvements on all five languages. 
This shows that (1) there is no clear winner between XLM-R and \Furina{} when applied on text written in different scripts, and (2) romanizing all languages did not improve zero-shot cross-lingual transfer for STR (\textbf{RQ3}). 

Next, we investigate the impact of augmenting the training data with translated data. %
The varied outcomes of augmenting data indicate that while machine translation can enhance transfer performance for certain languages. Performance drops in others may stem from shifts in label semantics and the degree of relatedness between original and translated sentence pairs (\textbf{RQ4}). Appendix~\ref{apdx:tq} (Table~\ref{fig:translation}) shows an example where MT fails to capture nuanced differences between closely, but not perfectly related sentences, leading to near-identical translations and inconsistent labels.

\paragraph{Single-source transfer results.}
We now select the most similar source languages based on different typological features obtained from the lang2vec (\texttt{L2V}) library. %
We obtain L2V vectors for Phonology (\texttt{Pho}), Syntax (\texttt{Syn}), Inventory (\texttt{Inv}), Family (\texttt{Fam}), Geography (\texttt{Geo}) and learned (\texttt{LRN}) features. 

Table~\ref{tab:results-l2v-ablation} shows our results for XLM-R.\footnote{\Furina{} results can be found in Appendix Table \ref{tab:results-l2v-single-furina}.}
Overall, a careful selection of a single-source language is crucial for zero-shot cross-lingual transfer. There is a substantial gap between the worst possible result (0.58) and the best possible result (0.73). On average, English is the most effective source language with a correlation of 0.71. A closer analysis reveals that English is the best language only for half of the target languages, despite being the language with the largest training dataset (cf. Table~\ref{app:l2vlangs} in Appendix). 
Interestingly, the best possible single-source language selection (MAX) results into the same performance as $\texttt{XLM-R}_\texttt{MS-All}$ (cf. Table~\ref{tab:results}).

\section{Conclusion}
In this paper, we investigate source language selection for cross-lingual transfer for Semantic Textual Relatedness (STR). We evaluate three different language selection strategies: single-source, multi-source transfer and transfer from English and two nearest language neighbors. 
We find that the transfer performance crucially depends on the size of the training dataset and the linguistic proximity to the test language. 
We further show that script differences cause high variance transfer performance and MT-based data augmentation can lead to shifts in label semantics. 
Fine-tuning \Furina{} on \texttt{eng}, \texttt{esp}, and \texttt{hau}, we achieve first place in the SemEval-2024 Task 1, Track C8 (\texttt{kin}). %

\section*{Acknowledgements}

This research is supported by the ERC Consolidator Grant DIALECT 101043235. 

\bibliography{custom,anthology}

\setlength{\tabcolsep}{6.5pt}
\begin{table*}%
\centering
\small 
\begin{tabular}{lllllll}
\toprule
Model Variant & Source languages & Target language & \# Train Instances & \Furina{} & XLM-R \\
\midrule
\multicolumn{6}{l}{Based on cell state vectors (\texttt{kNN}) \citep{malaviya-etal-2017-learning}} \\
\midrule
\multirow{2}{*}{1} & \multirow{2}{*}{esp, kin} & afr & \multirow{2}{*}{7840}& 0.80 & 0.81 \\
 & & hau & & 0.63 & 0.62 \\ \cdashline{1-6}[0.5pt/1pt]\noalign{\vskip 0.51ex} 
\multirow{2}{*}{2} & \multirow{2}{*}{esp, hau} & ind & \multirow{2}{*}{8798} & 0.46 & 0.45 \\
 & & kin & & 0.68 & 0.41 \\ \cdashline{1-6}[0.5pt/1pt]\noalign{\vskip 0.51ex} 
\multirow{2}{*}{3} &\multirow{2}{*}{kin, hau} & amh & \multirow{2}{*}{8014} & 0.74 & 0.75 \\
 & & esp & & 0.59 & 0.59 \\\cdashline{1-6}[0.5pt/1pt]\noalign{\vskip 0.51ex} 
4 & amh, hau & ary & 8228 & 0.57 & 0.50 \\
 \cdashline{1-6}[0.5pt/1pt]\noalign{\vskip 0.51ex} 
5 & kin, amh & arb & 7270 & 0.44 & 0.57 \\
 \cdashline{1-6}[0.5pt/1pt]\noalign{\vskip 0.51ex} 
6 & amh, esp & hin & 8054 & 0.72 & 0.78 \\ \cdashline{1-6}[0.5pt/1pt]\noalign{\vskip 0.51ex} 
avg & - & - & - & 0.58 & 0.58 \\
\midrule
\multicolumn{6}{l}{Based on learned lang2vec vectors (\texttt{L2V-LRN}) \citep{littell-etal-2017-uriel}} \\
\midrule
\multirow{4}{*}{1} & \multirow{4}{*}{esp, kin} & afr & \multirow{4}{*}{7840} & 0.80 & 0.81 \\
 & & arb &  & 0.46 & 0.60 \\
 & & hau &  & 0.63 & 0.62 \\
 & & ind &  & 0.44 & 0.39 \\
 \cdashline{1-6}[0.5pt/1pt]\noalign{\vskip 0.51ex} 
2 & esp, hau & kin & 8798 & 0.68 & 0.41 \\ \cdashline{1-6}[0.5pt/1pt]\noalign{\vskip 0.51ex} 
\multirow{2}{*}{3} & \multirow{2}{*}{kin, hau} & amh & \multirow{2}{*}{8014} & 0.74 & 0.75 \\
 & & esp &  & 0.59 & 0.59 \\ \cdashline{1-6}[0.5pt/1pt]\noalign{\vskip 0.51ex} 
4 & amh, hau & hin & 8228 & 0.74 & 0.79 \\ \cdashline{1-6}[0.5pt/1pt]\noalign{\vskip 0.51ex} 
5  & kin, amh & ary & 7270 & 0.52 & 0.55 \\
 \cdashline{1-6}[0.5pt/1pt]\noalign{\vskip 0.51ex} 
avg & - & - & - & 0.59 & 0.60 \\
\midrule
\multicolumn{6}{l}{Based on language familis features (\texttt{MS-Fam})} \\
\midrule
1 & esp, mar, tel & eng & 3932 & 0.83 & 0.82 \\ \cdashline{1-6}[0.5pt/1pt]\noalign{\vskip 0.51ex} 
2 & eng, mar, tel & esp & 7370 & 0.72 & 0.71 \\ \cdashline{1-6}[0.5pt/1pt]\noalign{\vskip 0.51ex} 
\multirow{3}{*}{3} & \multirow{3}{*}{eng, esp, mar, tel} & afr & \multirow{3}{*}{9432} & 0.79 & 0.81 \\ 
 & & hin & & 0.77 & 0.82 \\
 & & pan & & 0.02 & -0.00 \\ \cdashline{1-6}[0.5pt/1pt]\noalign{\vskip 0.51ex} 
4 & arq, ary, hau & amh & 3921 & 0.66 & 0.69 \\ \cdashline{1-6}[0.5pt/1pt]\noalign{\vskip 0.51ex} 
5 & amh, arq, ary, hau & arb & 4913 & 0.42 & 0.44 \\ \cdashline{1-6}[0.5pt/1pt]\noalign{\vskip 0.51ex} 
6 & amh, ary, hau & arq & 3652 & 0.55 & 0.37 \\ \cdashline{1-6}[0.5pt/1pt]\noalign{\vskip 0.51ex} 
7 & amh, arq, hau & ary & 3989 & 0.82 & 0.83 \\ \cdashline{1-6}[0.5pt/1pt]\noalign{\vskip 0.51ex} 
8 & amh, arq, ary & hau & 3117 & 0.68 & 0.66 \\ 
\bottomrule
\end{tabular}
\caption{Model variants based on language vectors, language cell state vectors and language families. All variants include \texttt{eng} for training.}
\label{tab:model_variants}
\end{table*}

\appendix

\section{Hyperparameter}
\label{sec:Hyperparameter}

We employed identical hyperparameters across all variants of XLM-R and \Furina. We train our models for at most 30 epochs with a batch size of 32 and a learning rate of 2e-5 and use AdamW \citep{loshchilov2017decoupled} with a weight decay of 1e-3. We evaluate the dev set performance every 200 steps and stop early based on the spearman correlation on the validation set (patience counter: 8, patience threshold: 1e-4).

\section{Lang2vec Augmentation}
We computed cosine similarity of all train and test languages based on L2V language vectors shown in Table~\ref{apdx:knn_cosine}-\ref{apdx:l2v_geo}. As showed in Table \ref{tab:model_variants}, we selected model variants with five and six combinations of two closest donor languages based on cell state vectors and learned language vectors, respectively. The correlation scores of model variants on target test languages are presented in the table for comparison with overall best scores in our main results. Ideally, each model variant should achieve the best performance on corresponding target test language due to the linguistic similarities between donor languages and the test language. Indeed, the highest score achieved for each target language does not necessarily correspond precisely with the selection of model variants based on language similarity calculated from language dense vectors.

\begin{table*}
\centering
\begin{tabular}{l | l l l l l}
 & amh & ary & esp & hau & kin \\ \midrule
afr & 0.75 & 0.57 & \cellcolor[HTML]{F8DBDB} 0.83 & 0.79 & 0.82 \\
amh & - & 0.62 & 0.66 & 0.69 & \cellcolor[HTML]{F8DBDB} 0.71 \\
ary & \cellcolor[HTML]{F8DBDB} 0.62 & - & 0.54 & 0.61 & 0.49 \\
arb & 0.76 & 0.73 & 0.73 & 0.73 & \cellcolor[HTML]{F8DBDB} 0.79 \\
esp & 0.66 & 0.54 & - & 0.76 & \cellcolor[HTML]{F8DBDB} 0.82 \\
hau & 0.69 & 0.61 & 0.76 & - & \cellcolor[HTML]{F8DBDB} 0.84 \\
hin & \cellcolor[HTML]{F8DBDB} 0.80 & 0.73 & 0.74 & 0.72 & 0.71 \\
ind & 0.71 & 0.65 & \cellcolor[HTML]{F8DBDB} 0.83 & 0.76 & 0.76 \\
kin & 0.71 & 0.49 & 0.82 & \cellcolor[HTML]{F8DBDB} 0.84 & - \\
\end{tabular}
\caption{Cosine similarities between source languages (rows) and target languages (columns). Language vectors are obtained from lang2vec:  \textbf{\texttt{kNN}} (cell\_state vectors) \citep{malaviya-etal-2017-learning}. We exclude four languages for which we cannot obtain feature vectors: arq, mar, tel, eng.}
\label{apdx:knn_cosine}
\end{table*}

\begin{table*}
\centering
\begin{tabular}{l | l l l l l}
 & amh & ary & esp & hau & kin \\ \midrule
afr & 0.07 & -0.05 & \cellcolor[HTML]{F8DBDB}0.23 & 0.07 & 0.22 \\
amh & - & -0.01 & 0.00 & \cellcolor[HTML]{F8DBDB}0.07 & 0.05 \\
ary & -0.01 & - & -0.06 & -0.03 & \cellcolor[HTML]{F8DBDB}0.06 \\
arb & 0.07 & -0.05 & 0.13 & -0.03 & \cellcolor[HTML]{F8DBDB}0.11 \\
esp & 0.00 & -0.06 & - & 0.22 & \cellcolor[HTML]{F8DBDB}0.23 \\
hau & 0.07 & -0.03 &\cellcolor[HTML]{F8DBDB} 0.22 & - & 0.19 \\
hin & \cellcolor[HTML]{F8DBDB}0.13 & -0.01 & 0.06 & 0.07 & 0.06 \\
ind & 0.00 & 0.05 & \cellcolor[HTML]{F8DBDB}0.11 & 0.06 & 0.09 \\
kin & 0.05 & 0.06 & \cellcolor[HTML]{F8DBDB}0.23 & 0.19 & - \\
\end{tabular}
\caption{Cosine similarities between source languages (rows) and target languages (columns). Language vectors are obtained from lang2vec:  \texttt{L2V-LRN} \citep{littell-etal-2017-uriel}. We exclude four languages for which we cannot obtain \textbf{\texttt{L2V-LRN}} features: arq, mar, tel, eng.}
\end{table*}

\begin{table*}
\centering

\begin{tabular}{l | l l l l l l l l l}
 & amh & ary & esp & hau & kin & arq & mar & tel & eng \\ \midrule
afr & 0.86 & 0.70 & 0.76 & \cellcolor[HTML]{F8DBDB} 0.87 & 0.85 & 0.73 & 0.80 & 0.80 & 0.82 \\
amh & - & 0.73 & 0.80 & 0.82 & 0.78 & 0.76 & \cellcolor[HTML]{F8DBDB} 0.95 & 0.84 & 0.76 \\
ary & 0.73 & - & 0.73 & 0.67 & 0.73 & \cellcolor[HTML]{F8DBDB} 0.97 & 0.77 & 0.69 & 0.70 \\
arb & 0.85 & 0.90 & 0.76 & 0.77 & 0.76 & \cellcolor[HTML]{F8DBDB} 0.93 & 0.80 & 0.71 & 0.73 \\
esp & 0.80 & 0.73 & - & 0.73 & 0.78 & 0.76 & 0.84 & 0.74 & \cellcolor[HTML]{F8DBDB} 0.86 \\
hau & \cellcolor[HTML]{F8DBDB} 0.82 & 0.67 & 0.73 & - & 0.82 & 0.69 & 0.77 & 0.77 & 0.78 \\
hin & 0.82 & 0.75 & 0.82 & 0.75 & 0.82 & 0.77 & \cellcolor[HTML]{F8DBDB} 0.87 & \cellcolor[HTML]{F8DBDB} 0.87 & 0.78 \\
ind & 0.76 & 0.70 & 0.76 & 0.78 & 0.85 & 0.73 & 0.80 & 0.80 & \cellcolor[HTML]{F8DBDB} 0.91 \\
kin & 0.78 & 0.73 & 0.78 & 0.82 & - & 0.76 & 0.82 & 0.82 & \cellcolor[HTML]{F8DBDB} 0.85 \\
arq & 0.76 & \cellcolor[HTML]{F8DBDB} 0.97 & 0.76 & 0.69 & 0.76 & - & 0.80 & 0.71 & 0.73 \\
eng & 0.76 & 0.70 & \cellcolor[HTML]{F8DBDB} 0.86 & 0.78 & 0.85 & 0.73 & 0.80 & 0.80 & - \\
pan & 0.95 & 0.77 & 0.84 & 0.77 & 0.82 & 0.80 & \cellcolor[HTML]{F8DBDB} 1.00 & 0.89 & 0.80 \\

\end{tabular}
\caption{Cosine similarities between source languages (rows) and target languages (columns). Language vectors are obtained from lang2vec:   \textbf{\texttt{L2V-Phon}} \citep{littell-etal-2017-uriel}.}
\end{table*}

\begin{table*}
\centering

\begin{tabular}{l | l l l l l l l l l}
 & amh & ary & esp & hau & kin & arq & mar & tel & eng \\ \midrule
afr & 0.62 & 0.66 & 0.73 & 0.71 & 0.55 & 0.67 & 0.62 & 0.56 & \cellcolor[HTML]{F8DBDB} 0.85 \\
amh & - & 0.59 & 0.63 & 0.57 & 0.51 & 0.60 & 0.72 &\cellcolor[HTML]{F8DBDB} 0.77 & 0.59 \\
ary & 0.59 & - & 0.81 & 0.72 & 0.63 & \cellcolor[HTML]{F8DBDB} 0.93 & 0.50 & 0.48 & 0.73 \\
arb & 0.61 &\cellcolor[HTML]{F8DBDB} 0.87 & 0.75 & 0.64 & 0.64 & 0.85 & 0.49 & 0.50 & 0.64 \\
esp & 0.63 & 0.81 & - & 0.74 & 0.59 & 0.81 & 0.56 & 0.52 & \cellcolor[HTML]{F8DBDB} 0.82 \\
hau & 0.57 & 0.72 & 0.74 & - & 0.65 & \cellcolor[HTML]{F8DBDB} 0.78 & 0.52 & 0.34 & 0.75 \\
hin & 0.74 & 0.67 & 0.68 & 0.57 & 0.46 & 0.65 & \cellcolor[HTML]{F8DBDB} 0.83 & 0.78 & 0.62 \\
ind & 0.45 & 0.73 & 0.66 & 0.67 & 0.52 & \cellcolor[HTML]{F8DBDB} 0.74 & 0.36 & 0.32 & 0.73 \\
kin & 0.51 & 0.63 & 0.59 & \cellcolor[HTML]{F8DBDB} 0.65 & - & 0.64 & 0.39 & 0.38 & 0.49 \\
arq & 0.60 & \cellcolor[HTML]{F8DBDB} 0.93 & 0.81 & 0.78 & 0.64 & - & 0.49 & 0.47 & 0.74 \\
eng & 0.59 & 0.73 & \cellcolor[HTML]{F8DBDB} 0.82 & 0.75 & 0.49 & 0.74 & 0.56 & 0.52 & - \\
pan & 0.71 & 0.68 & 0.70 & 0.59 & 0.49 & 0.67 & \cellcolor[HTML]{F8DBDB} 0.79 & 0.75 & 0.61 \\

\end{tabular}
\caption{Cosine similarities between source languages (rows) and target languages (columns). Language vectors are obtained from lang2vec:  \textbf{\texttt{L2V-Syn}} \citep{littell-etal-2017-uriel}.}
\end{table*}

\begin{table*}
\centering
\begin{tabular}{l | l l l l l l l l l}
 & amh & ary & esp & hau & kin & arq & mar & tel & eng \\ \midrule
afr & 0.65 & 0.56 & 0.62 & 0.61 & \cellcolor[HTML]{F8DBDB}0.69 & 0.61 & 0.67 & 0.68 & 0.69 \\
amh & - & 0.76 & 0.74 & \cellcolor[HTML]{F8DBDB}0.83 & 0.80 & 0.73 & 0.73 & 0.64 & 0.70 \\
ary & 0.76 & - & 0.62 & 0.70 & 0.70 & \cellcolor[HTML]{F8DBDB}0.85 & 0.63 & 0.57 & 0.65 \\
arb & 0.72 & 0.83 & 0.65 & 0.70 & 0.71 &\cellcolor[HTML]{F8DBDB} 0.98 & 0.64 & 0.60 & 0.73 \\
esp & \cellcolor[HTML]{F8DBDB}0.74 & 0.62 & - & 0.67 & 0.68 & 0.64 & 0.66 & 0.66 & 0.64 \\
hau & \cellcolor[HTML]{F8DBDB}0.83 & 0.70 & 0.67 & - & 0.76 & 0.72 & 0.64 & 0.59 & 0.62 \\
hin & 0.66 & 0.69 & 0.57 & 0.62 & 0.69 & \cellcolor[HTML]{F8DBDB}0.77 & 0.72 & \cellcolor[HTML]{F8DBDB}0.77 & 0.71 \\
ind & \cellcolor[HTML]{F8DBDB}0.88 & 0.75 & 0.76 & 0.79 & 0.82 & 0.77 & 0.74 & 0.68 & 0.76 \\
kin & \cellcolor[HTML]{F8DBDB}0.80 & 0.70 & 0.68 & 0.76 & - & 0.72 & 0.65 & 0.63 & 0.69 \\
arq & 0.73 &\cellcolor[HTML]{F8DBDB} 0.85 & 0.64 & 0.72 & 0.72 & - & 0.65 & 0.62 & 0.71 \\
eng & 0.70 & 0.65 & 0.64 & 0.62 & 0.69 & 0.71 &\cellcolor[HTML]{F8DBDB} 0.76 & 0.67 & - \\
pan & 0.71 & 0.60 & 0.69 & 0.65 & 0.71 & 0.66 & \cellcolor[HTML]{F8DBDB}0.82 & 0.78 & 0.77 \\
\end{tabular}
\caption{Cosine similarities between source languages (rows) and target languages (columns). Language vectors are obtained from lang2vec:  \textbf{\texttt{L2V-Inv}} \citep{littell-etal-2017-uriel}.}
\end{table*}

\begin{table*}
\centering
\begin{tabular}{l | l l l l l l l l l}
 & amh & ary & esp & hau & kin & arq & mar & tel & eng \\ \midrule
afr & 0.00 & 0.00 & 0.11 & 0.00 & 0.00 & 0.00 & 0.15 & 0.00 & \cellcolor[HTML]{F8DBDB} 0.50 \\
amh & - & 0.40 & 0.00 & 0.17 & 0.00 & \cellcolor[HTML]{F8DBDB} 0.43 & 0.00 & 0.00 & 0.00 \\
ary & 0.40 & - & 0.00 & 0.16 & 0.00 & \cellcolor[HTML]{F8DBDB} 0.94 & 0.00 & 0.00 & 0.00 \\
arb & 0.46 & 0.87 & 0.00 & 0.18 & 0.00 & \cellcolor[HTML]{F8DBDB} 0.93 & 0.00 & 0.00 & 0.00 \\
esp & 0.00 & 0.00 & - & 0.00 & 0.00 & 0.00 & \cellcolor[HTML]{F8DBDB} 0.12 & 0.00 & 0.10 \\
hau & \cellcolor[HTML]{F8DBDB} 0.17 & 0.16 & 0.00 & - & 0.00 & \cellcolor[HTML]{F8DBDB} 0.17 & 0.00 & 0.00 & 0.00 \\
hin & 0.00 & 0.00 & 0.11 & 0.00 & 0.00 & 0.00 &\cellcolor[HTML]{F8DBDB} 0.46 & 0.00 & 0.13 \\
ind & 0.00 & 0.00 & 0.00 & 0.00 & 0.00 & 0.00 & 0.00 & 0.00 & 0.00 \\
kin & 0.00 & 0.00 & 0.00 & 0.00 & - & 0.00 & 0.00 & 0.00 & 0.00 \\
arq & 0.43 & \cellcolor[HTML]{F8DBDB} 0.94 & 0.00 & 0.17 & 0.00 & - & 0.00 & 0.00 & 0.00 \\
eng & 0.00 & 0.00 & 0.10 & 0.00 & 0.00 & 0.00 & \cellcolor[HTML]{F8DBDB} 0.14 & 0.00 & - \\
pan & 0.00 & 0.00 & 0.12 & 0.00 & 0.00 & 0.00 & \cellcolor[HTML]{F8DBDB} 0.50 & 0.00 & 0.14 \\
\end{tabular}
\caption{Cosine similarities between source languages (rows) and target languages (columns). Language vectors are obtained from lang2vec: \textbf{\texttt{L2V-Fam}} \citep{littell-etal-2017-uriel}.}
\end{table*}

\begin{table*}
\centering

\begin{tabular}{l | l l l l l l l l l}
 & amh & ary & esp & hau & kin & arq & mar & tel & eng \\ \midrule
afr & 0.97 & 0.91 & 0.90 & 0.96 & \cellcolor[HTML]{F8DBDB}0.99 & 0.91 & 0.92 & 0.92 & 0.87 \\
amh & - & 0.95 & 0.95 & 0.98 & \cellcolor[HTML]{F8DBDB}0.99 & 0.96 & 0.97 & 0.96 & 0.94 \\
ary & 0.95 & - & \cellcolor[HTML]{F8DBDB}1.00 & 0.98 & 0.94 & \cellcolor[HTML]{F8DBDB}1.00 & 0.88 & 0.87 & 0.99 \\
arb & \cellcolor[HTML]{F8DBDB}0.99 & 0.95 & 0.96 & 0.97 & 0.97 & 0.97 & 0.98 & 0.97 & 0.96 \\
esp & 0.95 & \cellcolor[HTML]{F8DBDB}1.00 & - & 0.98 & 0.94 & \cellcolor[HTML]{F8DBDB}1.00 & 0.90 & 0.89 & \cellcolor[HTML]{F8DBDB}1.00 \\
hau & 0.98 & 0.98 & 0.98 & - & \cellcolor[HTML]{F8DBDB}0.99 & 0.98 & 0.90 & 0.90 & 0.96 \\
hin & 0.97 & 0.89 & 0.91 & 0.90 & 0.94 & 0.91 & \cellcolor[HTML]{F8DBDB}1.00 & \cellcolor[HTML]{F8DBDB}1.00 & 0.91 \\
ind & 0.89 & 0.77 & 0.79 & 0.81 & 0.87 & 0.79 & \cellcolor[HTML]{F8DBDB}0.96 & \cellcolor[HTML]{F8DBDB}0.96 & 0.79 \\
kin & \cellcolor[HTML]{F8DBDB}0.99 & 0.94 & 0.94 & \cellcolor[HTML]{F8DBDB}0.99 & - & 0.95 & 0.94 & 0.94 & 0.92 \\
arq & 0.96 & \cellcolor[HTML]{F8DBDB}1.00 & \cellcolor[HTML]{F8DBDB}1.00 & 0.98 & 0.95 & - & 0.90 & 0.90 & 0.99 \\
eng & 0.94 & 0.99 & \cellcolor[HTML]{F8DBDB}1.00 & 0.96 & 0.92 & 0.99 & 0.90 & 0.89 & - \\
pan & 0.96 & 0.90 & 0.91 & 0.91 & 0.93 & 0.92 & \cellcolor[HTML]{F8DBDB}1.00 & \cellcolor[HTML]{F8DBDB}1.00 & 0.92 \\

\end{tabular}
\caption{Cosine similarities between source languages (rows) and target languages (columns). Language vectors are obtained from lang2vec: \textbf{\texttt{L2V-Geo}} \citep{littell-etal-2017-uriel}. }
\label{apdx:l2v_geo}
\end{table*}

\setlength{\tabcolsep}{7.3pt}
\begin{table*}[t!]
\centering
\small 
\begin{tabular}{l c c c c c c c c c c c c l } \toprule
& \multicolumn{5}{c}{Indo-European} & \multicolumn{5}{c}{Afro-Asiatic} & \multicolumn{2}{c}{Other} \\ \cmidrule(lr){2-6} \cmidrule(lr){7-11} \cmidrule(lr){12-13}
\textbf{\Furina}
 & eng & esp & afr & hin & pan & amh & arb & arq & ary & hau & ind & kin & avg \\ \midrule
MIN & 0.34 & 0.38 & 0.48 & 0.35 & -0.19 & 0.68 & 0.04 & 0.00 & 0.28 & 0.33 & 0.22 & 0.23 & 0.36 \\ \midrule
$\texttt{Furina}_\texttt{eng}$ & - & 0.54 & 0.79 & 0.70 & -0.14 & 0.74 & 0.37 & 0.45 & 0.59 & 0.63 & 0.44 & 0.53 & 0.62 \\ 
$\texttt{Furina}_\texttt{L2V-kNN}$ & - & 0.62 & 0.71 & 0.35 & - & 0.73 & 0.42 & - & 0.28 & 0.64 & 0.42 & 0.68 & 0.53 \\ \cdashline{1-14}[.4pt/1pt]\noalign{\vskip 0.51ex}
\texttt{L2V-Pho} & 0.76 & 0.56 & 0.79 & 0.77 & 0.03 & 0.76 & 0.46 & 0.48 & 0.63 & 0.43 & 0.43 & 0.68 & 0.63 \\
\texttt{L2V-Syn} & 0.76 & 0.56 & 0.80 & 0.78 & 0.03 & 0.74 & 0.39 & 0.48 & 0.63 & 0.54 & 0.34 & 0.68 & 0.63 \\
\texttt{L2V-Inv} & 0.78 & 0.38 & 0.79 & 0.76 & 0.03 & 0.76 & 0.46 & 0.48 & 0.63 & 0.43 & 0.22 & 0.23 & 0.60 \\
\texttt{L2V-Fam} & 0.78 & 0.64 & 0.80 & 0.78 & 0.03 & 0.76 & 0.46 & 0.48 & 0.63 & 0.485 & - & - & 0.65 \\
\texttt{L2V-Geo} & 0.76 & 0.47 & 0.79 & 0.78 & 0.03 & 0.73 & 0.04 & 0.53 & 0.63 & 0.64 & 0.30 & 0.23 & 0.58 \\ %
\texttt{L2V-LRN} & - & 0.62 & 0.71 & 0.35 & - & 0.76 & 0.42 & - & 0.59 & 0.33 & 0.42 & 0.54 & 0.54 \\ \midrule
MAX & 0.79 & 0.64 & 0.81 & 0.78 & 0.06 & 0.76 & 0.53 & 0.55 & 0.77 & 0.66 & 0.45 & 0.78 & 0.71 \\  \bottomrule

\end{tabular}
\caption{
Single-source transfer results in terms of spearman correlation. The language selection is based on the cosine similarity of different typological features obtained from lang2vec (\texttt{L2V}). We additionally report the lower and upper bound (MIN, MAX) when choosing the worst and best possible donor language for each test language. Languages that are not covered by all L2V features are excluded from the average (\texttt{eng}, \texttt{pan}, \texttt{arq}, \texttt{ind}, \texttt{kin}). 
}
\label{tab:results-l2v-single-furina}
\end{table*}

\begin{table*}
\centering
\small  

\begin{tabular}{l l l l l l l l l l l l l} \toprule
 & afr & amh & ary & arb & esp & hau & hin & ind & kin & arq & eng & pan \\ \midrule
MIN (\texttt{XLM-R}) & \cellcolor[HTML]{FFD966}esp &
  \cellcolor[HTML]{FFD966}esp &
  \cellcolor[HTML]{D5A6BD}amh &
  \cellcolor[HTML]{D5A6BD}amh &
  \cellcolor[HTML]{B6D7A8}ary &
  \cellcolor[HTML]{FFD966}esp &
  \cellcolor[HTML]{FFD966}esp &
  \cellcolor[HTML]{C27BA0}tel &
  \cellcolor[HTML]{FFE599}arq &
  \cellcolor[HTML]{D5A6BD}amh &
  \cellcolor[HTML]{FFD966}esp &
  \cellcolor[HTML]{B6D7A8}ary \\
MIN (\texttt{Furina}) &   
  \cellcolor[HTML]{D5A6BD}amh &
  \cellcolor[HTML]{FFD966}esp &
  \cellcolor[HTML]{D5A6BD}amh &
  \cellcolor[HTML]{D5A6BD}amh &
  \cellcolor[HTML]{D5A6BD}amh &
  \cellcolor[HTML]{FFD966}esp &
  \cellcolor[HTML]{D5A6BD}amh &
  \cellcolor[HTML]{D5A6BD}amh &
  \cellcolor[HTML]{D5A6BD}amh &
  \cellcolor[HTML]{D5A6BD}amh &
  \cellcolor[HTML]{D5A6BD}amh &
  \cellcolor[HTML]{B6D7A8}ary \\ \midrule%
\texttt{kNN} & 
  \cellcolor[HTML]{FFD966}esp &
  \cellcolor[HTML]{FF9900}kin &
  \cellcolor[HTML]{D5A6BD}amh &
  \cellcolor[HTML]{FF9900}kin &
  \cellcolor[HTML]{FF9900}kin & 
  \cellcolor[HTML]{FF9900}kin &
  \cellcolor[HTML]{D5A6BD}amh &
  \cellcolor[HTML]{FFD966}esp &
  \cellcolor[HTML]{A4C2F4}hau &
 - & - & - \\
\texttt{L2V-Pho} &
  \cellcolor[HTML]{A4C2F4}hau &
  \cellcolor[HTML]{93C47D}mar &
  \cellcolor[HTML]{FFE599}arq &
  \cellcolor[HTML]{FFE599}arq &
  \cellcolor[HTML]{E06666}eng &
  \cellcolor[HTML]{D5A6BD}amh &
  \cellcolor[HTML]{38FFF8}mar+tel &
  \cellcolor[HTML]{E06666}eng &
  \cellcolor[HTML]{E06666}eng &
  \cellcolor[HTML]{B6D7A8}ary &
  \cellcolor[HTML]{FFD966}esp &
  \cellcolor[HTML]{93C47D}mar \\
\texttt{L2V-Syn} &
  \cellcolor[HTML]{E06666}eng &
  \cellcolor[HTML]{C27BA0}tel &
  \cellcolor[HTML]{FFE599}arq &
  \cellcolor[HTML]{B6D7A8}ary &
  \cellcolor[HTML]{E06666}eng &
  \cellcolor[HTML]{FFE599}arq &
  \cellcolor[HTML]{93C47D}mar &
  \cellcolor[HTML]{FFE599}arq &
  \cellcolor[HTML]{A4C2F4}hau &
  \cellcolor[HTML]{B6D7A8}ary &
  \cellcolor[HTML]{FFD966}esp &
  \cellcolor[HTML]{93C47D}mar \\
\texttt{L2V-Inv} &
  \cellcolor[HTML]{FF9900}kin &
  \cellcolor[HTML]{A4C2F4}hau &
  \cellcolor[HTML]{FFE599}arq &
  \cellcolor[HTML]{FFE599}arq &
  \cellcolor[HTML]{D5A6BD}amh &
  \cellcolor[HTML]{D5A6BD}amh &
  \cellcolor[HTML]{C27BA0}tel &
  \cellcolor[HTML]{D5A6BD}amh &
  \cellcolor[HTML]{D5A6BD}amh &
  \cellcolor[HTML]{B6D7A8}ary &
  \cellcolor[HTML]{93C47D}mar &
  \cellcolor[HTML]{93C47D}mar \\
\texttt{L2V-Fam} &
  \cellcolor[HTML]{E06666}eng &
  \cellcolor[HTML]{FFE599}arq &
  \cellcolor[HTML]{FFE599}arq &
  \cellcolor[HTML]{FFE599}arq &
  \cellcolor[HTML]{93C47D}mar &
  \cellcolor[HTML]{FCFF2F}amh+arq &
  \cellcolor[HTML]{93C47D}mar &
  - &
  - &
  \cellcolor[HTML]{B6D7A8}ary &
  \cellcolor[HTML]{93C47D}mar &
  \cellcolor[HTML]{93C47D}mar \\
\texttt{L2V-Geo} &
  \cellcolor[HTML]{FF9900}kin &
  \cellcolor[HTML]{FF9900}kin &
  \cellcolor[HTML]{FFE599}arq &
  \cellcolor[HTML]{D5A6BD}amh &
  \cellcolor[HTML]{FFE599}arq &
  \cellcolor[HTML]{FF9900}kin &
  \cellcolor[HTML]{93C47D}mar &
  \cellcolor[HTML]{C27BA0}tel &
  \cellcolor[HTML]{D5A6BD}amh &
  \cellcolor[HTML]{FFD966}esp &
  \cellcolor[HTML]{FFD966}esp &
  \cellcolor[HTML]{93C47D}mar \\
\texttt{L2V-LRN} & 
  \cellcolor[HTML]{FFD966}esp &
  \cellcolor[HTML]{A4C2F4}hau &
  \cellcolor[HTML]{FF9900}kin &
  \cellcolor[HTML]{FF9900}kin &
  \cellcolor[HTML]{FF9900}kin &
  \cellcolor[HTML]{FFD966}esp &
  \cellcolor[HTML]{D5A6BD}amh &
  \cellcolor[HTML]{FFD966}esp &
  \cellcolor[HTML]{FFD966}esp &
  - &
  - &
  - \\ \midrule
MAX (\texttt{XLM-R}) &
  \cellcolor[HTML]{E06666}eng &
  \cellcolor[HTML]{E06666}eng &
  \cellcolor[HTML]{E06666}eng &
  \cellcolor[HTML]{93C47D}mar &
  \cellcolor[HTML]{A4C2F4}hau &
  \cellcolor[HTML]{E06666}eng &
  \cellcolor[HTML]{E06666}eng &
  \cellcolor[HTML]{FFD966}esp &
  \cellcolor[HTML]{93C47D}mar &
  \cellcolor[HTML]{E06666}eng &
  \cellcolor[HTML]{93C47D}mar &
  \cellcolor[HTML]{D5A6BD}amh \\
MAX (\texttt{Furina}) &
  \cellcolor[HTML]{93C47D}mar &
  \cellcolor[HTML]{FFE599}arq &
  \cellcolor[HTML]{E06666}eng &
  \cellcolor[HTML]{E06666}eng &
  \cellcolor[HTML]{93C47D}mar &
  \cellcolor[HTML]{93C47D}mar &
  \cellcolor[HTML]{93C47D}mar &
  \cellcolor[HTML]{B6D7A8}ary &
  \cellcolor[HTML]{93C47D}mar &
  \cellcolor[HTML]{93C47D}mar &
  \cellcolor[HTML]{A4C2F4}hau &
  \cellcolor[HTML]{FF9900}kin \\\bottomrule
\end{tabular}
\caption{Each cell shows a given test language and lang2vec (\texttt{L2V}) feature the closest source language used for single source transfer in Table~\ref{tab:results-l2v-ablation} and Table~\ref{tab:results-l2v-single-furina}. We further show the closest languages according to cell-state vectors obtained from a multilingual MT system (\texttt{kNN}) \citep{malaviya-etal-2017-learning}, see \S\ref{sec:da} for details. MIN and MAX show the source language for which best transfer and worst transfer performance is achieved. }
\label{app:l2vlangs}
\end{table*}

\section{Translation quality.} 
\label{apdx:tq}

\begin{table}[h!]
\begin{small}
    \centering
    \begin{tabular}{p{1.5cm}p{5.5cm}}
    \textbf{Pair} & \textbf{Sentence Pair}\\
    \toprule
    \multirow{ 2}{*}{esp-182}  & ``Un hombre está saltando a una pared baja.''\\
   & 
``Un hombre está saltando a un muro bajo.'' \\
\midrule
    \multirow{ 2}{*}{translated}  & ``A man is jumping into a low wall.'' \\
    & ``A man is jumping into a low wall''\\
    \bottomrule
    \end{tabular}
    \end{small}
    \caption{An example from Spanish training dataset with its English translation, the label is 0.80.}
    \label{fig:translation}
\end{table}

We reviewed some machine-translated examples and noticed that subtle differences in the original language can be lost during translation. As shown in Table~\ref{fig:translation}, the two translated sentences, apart from punctuation, share no differences while the label assigned is 0.8. This undoubtedly has the potential to interfere with the model's learning process for the STR task.

\end{document}